\documentclass[journal]{IEEEtran}

\usepackage{float} \restylefloat{figure} 
\usepackage[usenames]{color}

\usepackage{graphicx}
\usepackage{rotating}
\usepackage{lscape}
\usepackage{bm} 
\usepackage{cite}
\usepackage{url}
\usepackage{threeparttable}
\usepackage{tabularx}
\usepackage[small]{caption}
\usepackage{subfig} 
\usepackage{array} 
\usepackage{epstopdf}
\usepackage{amsmath}   
\usepackage{afterpage}
\begin{document}

\title{A 128 channel Extreme Learning Machine based Neural Decoder for Brain Machine Interfaces}


\author{Yi Chen,~\IEEEmembership{Student Member,~IEEE}, Enyi Yao,~\IEEEmembership{Student Member,~IEEE}, Arindam Basu,~\IEEEmembership{Member,~IEEE}
\thanks{Yi Chen, Enyi Yao, and Arindam Basu are with Centre of Excellence in IC Design (VIRTUS), School of Electrical and Electronic Engineering,
Nanyang Technological University, Singapore 639798 (e-mail:ychen3@ntu.edu.sg, arindam.basu@ntu.edu.sg).}
}
\maketitle

\begin{abstract}

Currently, state-of-the-art motor intention decoding algorithms in brain-machine interfaces are mostly implemented on a PC and consume significant amount of power. A machine learning co-processor in 0.35-$\mu$m CMOS for the motor intention decoding in the brain-machine interfaces is presented in this paper. Using Extreme Learning Machine algorithm and low-power analog processing, it achieves an energy efficiency of $3.45$ pJ/MAC at a classification rate of $50$ Hz. The learning in second stage and corresponding digitally stored coefficients are used to increase robustness of the core analog processor. The chip is verified with neural data recorded in monkey finger movements experiment, achieving a decoding accuracy of $99.3\%$ for movement type. The same co-processor is also used to decode time of movement from asynchronous neural spikes. With time-delayed feature dimension enhancement, the classification accuracy can be increased by $5\%$ with limited number of input channels. Further, a sparsity promoting training scheme enables reduction of number of programmable weights by $\approx 2X$. 

\end{abstract}


\IEEEpeerreviewmaketitle


\section{Introduction}
\label{sec:intro}
\begin{figure}[!t]
\centerline{
\includegraphics[width=0.4\textwidth]{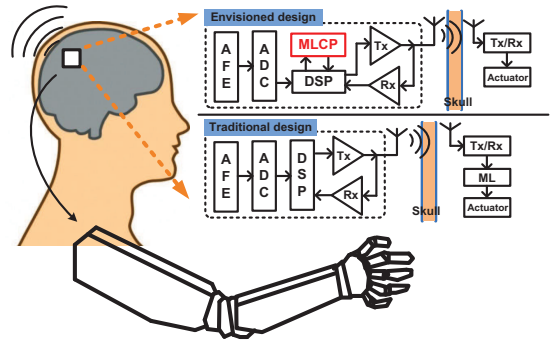}
}  \caption[Comparison of envisioned and traditional implanted BMI]{{\bf Comparison of envisioned and traditional implanted BMI:} The envisioned system uses a machine learning co-processor (MLCP) along with the DSP used in traditional neural implants to estimate motor intentions from neural recordings thus providing data compression. Traditional systems perform such decoding outside the implant and use bulky computers.} \label{fig:BMI}
\end{figure}

Brain-machine interfaces (BMI) are becoming increasingly popular over the last decade and open up the possibility of neural prosthetic devices for patients with paralysis or in locked-in state. As depicted in Fig. \ref{fig:BMI}, a typical implanted BMI consists of a neural recording IC to amplify, digitize and transmit neural action potentials (AP) recorded by the micro-electrode array (MEA). Significant effort has been dedicated to develop energy efficient neural recording channel in recent years for long-term operation of the implanted devices\cite{Harrison2007_sys}\cite{Sarpeshkar2008_BMI}\cite{Shahrokhi2010}\cite{Chen2014}. Some recent solutions have also integrated AP detection\cite{enyi_spd_iscas}\cite{Holleman2008}\cite{spd_yangzhi}\cite{spd_gosselin} and spike sorting features\cite{Chen2009a}\cite{karkare_2011}\cite{karkare_2013}. However, in order to produce an actuation command (e.g. for a prosthetic arm), the subsequent step of motor intention decoding is required to map spike train patterns acquired in the neural recording to the motor intention of the subjects. 

Though various elaborate models and methods of motor intention decoding have been developed in past decades with the goal of achieving high decoding performance \cite{Acharya2008}\cite{Ifft2013}\cite{Hochberg2012}, the state-of-the art neural signal decoding are mainly conducted on PC consuming a considerable amount of power and making it impractical for the long-term use. With on-chip real-time motor intention decoding, the size and the power consumption of the computing device can be reduced effectively and the solution becomes truly portable. Furthermore, integrating the neural decoding algorithm with the neural recording device is also desired to reduce the wireless data transmission rate and make the implanted BMI solution scalable as required in the future\cite{neural_rec_scaling}. Until now, very few attempts have been made to give a solution for this problem. A low-power motor intention architecture using analog computing is proposed in \cite{Rapoport2009}, featuring an active filtering with massive parallel computing through low power analog filters and memories. However, no measurement results are published to support the silicon viability of the architecture. A more recent work proposes a universal computing architecture for the neural signal decoding \cite{Rapoport2012}. The architecture is implemented on a FPGA with a power consumption of $537$ $\mu$W.



In this paper, we present a machine learning co-processor (MLCP) achieving low-power operation through massive parallelism, sub-threshold analog processing and careful choice of algorithm. Figure \ref{fig:BMI} contrasts our approach with traditional approaches: our MLCP acts in conjunction with the digital signal processor (DSP) already present in implants (for spike sorting, detection and packetizing) to provide the decoded outputs. The bulk of processing is done on the MLCP while simple digital functions are performed on the DSP. Compared to traditional designs that perform the decoding outside the implant, our envisioned system that provides opportunity for huge data compression by integrating the decoder in the implant. The MLCP is characterized by measurement and the decoding performance of the proposed design is verified with data acquired in individuate finger movements experiment of monkeys. Some initial results of this work were presented in \cite{yi_elm_iscas}. Here, we present more detailed theory, experimental results including decoding time of movement, new sparsity promoting training and also discuss scalability of this architecture.

\begin{figure}[!t]
	\centerline{
		\includegraphics[width=0.48\textwidth]{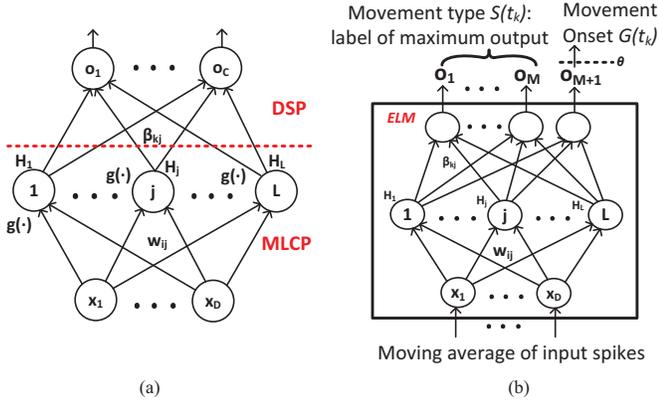}}

	\caption[Algorithm]{{\bf Algorithm:} (a) The architecture of the Extreme Learning Machine (ELM) with one nonlinear hidden layer and linear output layer. (b) Use of ELM in neural decoding for classifying movement type and onset time of movement.}
	\label{fig:ELM}
\end{figure}

\begin{figure*}[t]
	\centerline{
		\includegraphics[width=0.6\textwidth]{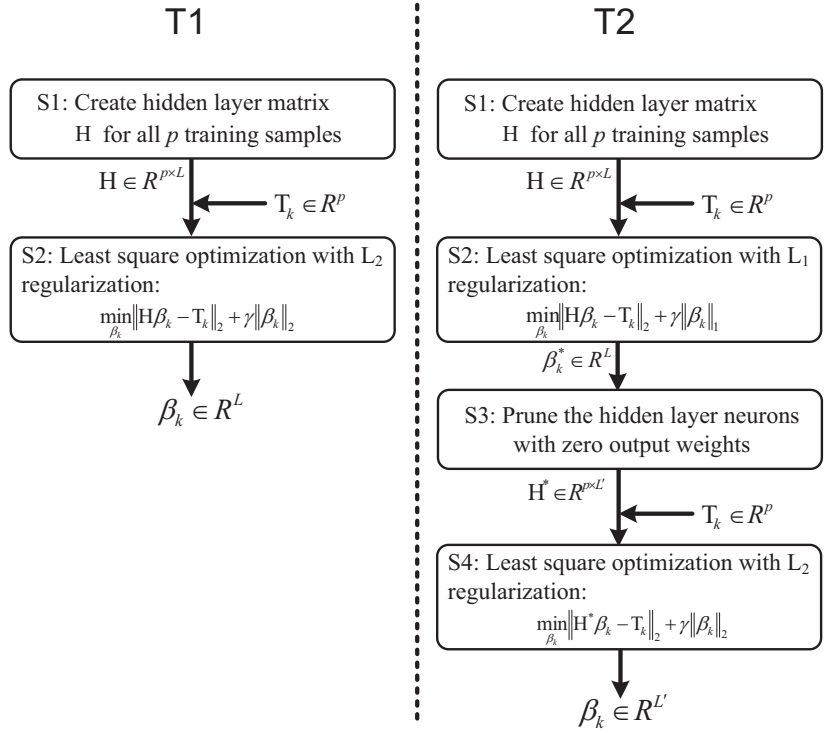}
	}  \caption[Training methods for ELM]{{\bf Training methods for ELM:} T1 is the conventionally used training method to improve generalization by minimizing norm of weights as well as training error. T2 uses an additional step of sparsifying output weights to reduce the required hardware.} \label{fig:training}
\end{figure*}

\section{Proposed Design: Algorithm}
\label{sec:algo}
\subsection{Extreme Learning Machine}
\label{sec:elm}
\subsubsection{Network Architecture}
The machine learning algorithm used in this design is Extreme Learning Machine (ELM) proposed in \cite{elm_neurocomp}. As depicted in Fig. \ref{fig:ELM} (a), The ELM is essentially a single hidden-layer feed-forward network (SLFN). The k-th output of the network (1$\leq$ k $\leq$ C) can be expressed as follows,
\begin{equation}
\begin{split}
o_{k}=\sum_{i}^{L}\beta_{ki}g\left ( \mathbf{w_ {i}},\mathbf{x},b_{i}\right )=\sum_{i}^{L}\beta_{ki}h_i=\mathbf{h}^T\boldsymbol{\beta}_k, \\ \mathbf{w_{i}}, \mathbf{x} \in R^{D}; \beta_{ki}, b_{i} \in R; \mathbf{h}, \boldsymbol{\beta}_k \in R^L 
\end{split}
\label{eq:ELM_out}
\end{equation}
where $\bf{x}$ denotes the input feature vector, $L$ is the number of hidden neurons, $\bf{h}$ is the output of the hidden layer, $b_{i}$ is the bias for each hidden layer node, $\bf{w_i}$ and $\beta_{ki}$ are input and output weights respectively. A non-linear activation function $g()$ is needed for non-linear classification. A special case of the nonlinear function is the additive node defined by $h_i=g\left ( \mathbf{w_ {i}^{T}}\mathbf{x}+b_{i}\right )$. The above equation can be compactly written for all classes as $o=\mathbf{h}\beta ,o=[o_1,o_2,..o_c]$
%
where $\beta=[\boldsymbol{\beta}_1,\boldsymbol{\beta}_2...\boldsymbol{\beta}_C]$ denotes the $L\times C$ matrix of output weights.

While the output $o_k$ can be directly used in regression, for classification tasks the input is categorized as the k-th class if $o_{k}$ is the largest output. Formally, we can define the classification output as an integer class label $s$ given by $s=argmax_k o_k,1\leq k\leq C$. Intuitively, we can think of the first layer as creating a set of random basis functions while the second layer chooses how to combine these functions to match a desired target. Of course, if we could choose the basis functions through training as well, we would need less number of such functions. But the penalty to be paid is longer training times. More details about the algorithm can be found in \cite{elm_neurocomp,Huang2012}.

\begin{figure*}[t]
\centerline{
\includegraphics[width=0.6\textwidth]{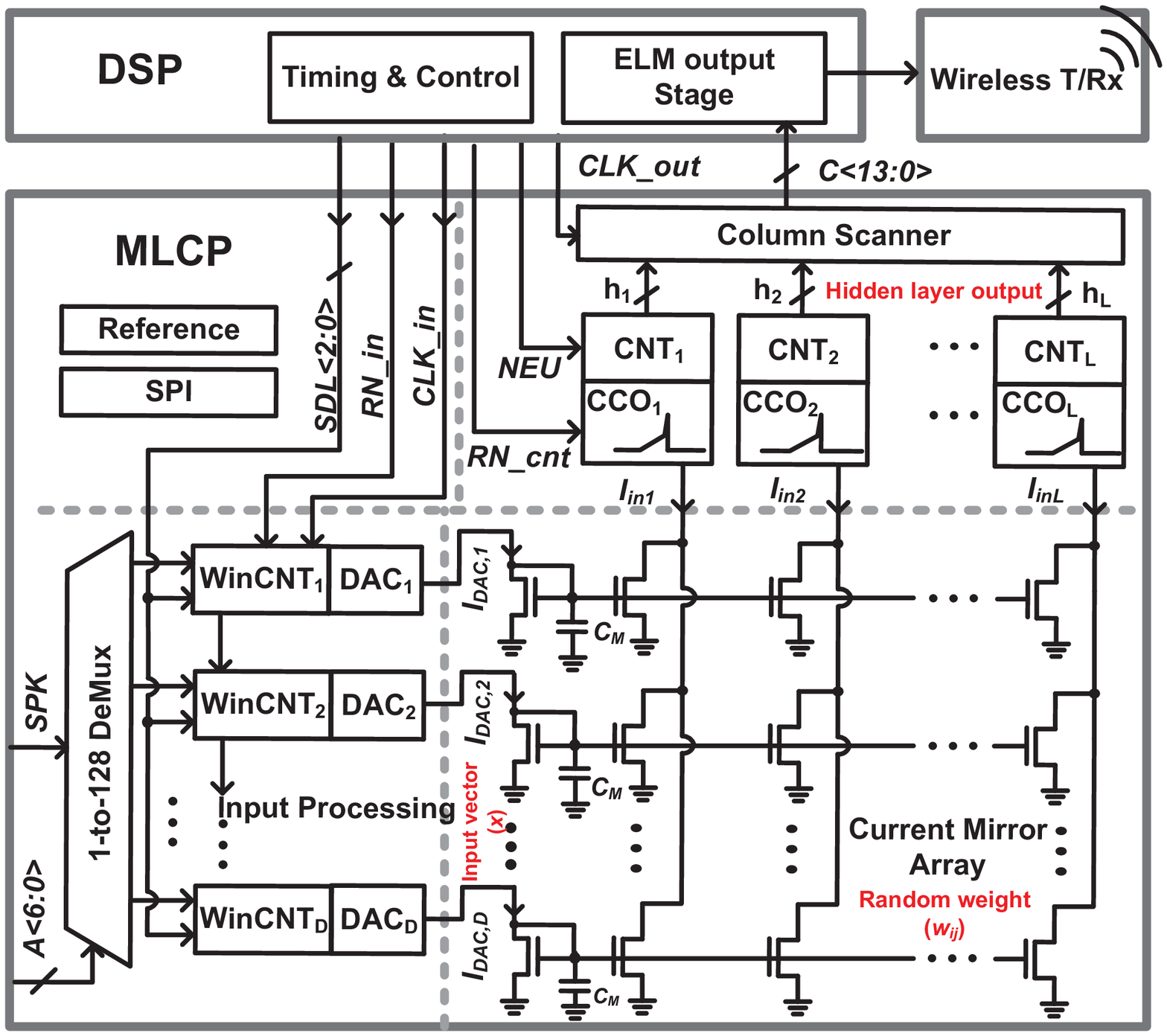}}
\centerline{(a)}

\centerline{
\includegraphics[width=0.6\textwidth]{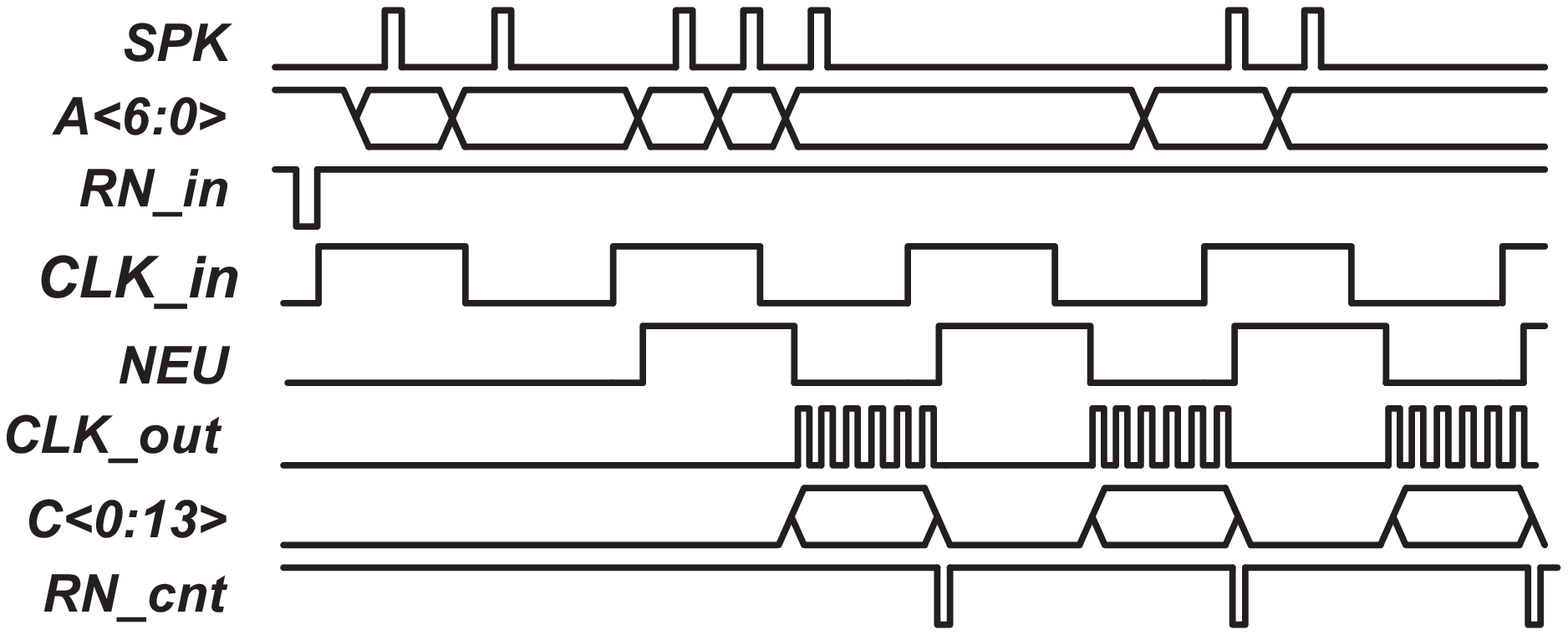}}
\centerline{(b)}

\caption[The diagram and the timing of MLCP-based decoding PEU]{{\bf The diagram (a) and the timing (b) of MLCP based neural decoder.}}
\label{fig:diagram}
\end{figure*}

\subsubsection{Training Methods}
\label{sec:training}
The special property of the ELM is that $\bf{w}$ can be random numbers from any continuous probability distribution and remains unchanged after initiation\cite{elm_neurocomp}, while only $\beta$ needs to be trained and stored with high resolution. Therefore the training of this SLFN reduces to finding a least-square solution of $\beta$ given the desired target values in a training set. We will next show two methods of training--the conventional one (T1) for improved generalization as well as a second method (T2) that promotes sparsity. For simplicity, we show the solution of weights for one output $o_k$--the same method can be extended to other output weights as well and can be represented in a compact matrix equation\cite{elm_neurocomp}. 

Suppose there are $p$ training samples--then we can create a $p\times L$ hidden layer output matrix $\bf{H}$ where each row of $\bf{H}$ has the hidden layer neuron output for each training sample. Let $\textbf{T}_k\in R^p$ be the vector of target values for the $p$ samples. With these inputs, the two training methods are shown in Fig. \ref{fig:training}. The step for $L_2$ norm minimization can be solved directly with the solution given by
$\boldsymbol{\beta}_k=\mathbf{H}^\dagger \mathbf{T}_k$ where $\bf{H}^\dagger$ is the Moore-Penrose generalized inverse of the matrix $\bf{H}$. Hence, training can happen quickly in this case. The $L_1$ norm minimization step in T2 however has to be performed using standard optimization algorithms like LARS\cite{lars}. Thus T2 provides reduced hardware complexity due to reduction in the number of hidden neurons at the cost of increased training time.

\subsection{Neural Decoding}
\label{sec:decoding}
The neural decoding algorithm we use is inspired by the method in \cite{Aggarwal2008}. We replace the committee of ANN in their work with ELM in our case. Three specific advantages of the ELM for this application are (1) the fixed random input weights can be realized by a current mirror array exploiting fabrication mismatch of the CMOS process; (2) one-step training that is necessary for quick weight update to address change in input statistics and (3) the hidden layer outputs $\bf{h}$ can be reused for multiple operations on the same input data $\bf{x}$. In this case, we have reused $\bf{h}$ to classify both the onset time and type of movement. One disadvantage of the ELM algorithm is the usage of $1.5-3X$ hidden neurons compared to fully tuned architectures (e.g. SVM, AdaBoost) since the hidden nodes in ELM only create random projections that are not fine tuned\cite{kitchen-sink}. However, implementing random weights results in more than $10X$ savings over fully tunable weights making this architecture more lucrative overall. Next, we give an overview of the decoding algorithm while the reader is pointed to \cite{Aggarwal2008} for more details.

\subsubsection{Movement type and Onset time Decoding}
Figure \ref{fig:ELM} (b) depicts how the ELM is used in neural decoding. Even though the input is an asynchronous spike train, the ELM produces classification outputs at a fixed rate of once every $T_s$ seconds. The input $\bf{x}$ is created from the firing rate of spike trains $p(t)=\sum_{t_s}\delta (t-t_s)$ of biological neurons by finding the moving average over a duration $T_w$. Hence, we can define the firing rate $r_i$ of i-th neuron at time instant $t_k$ as $r_i(t_k)=\int_{t_k-T_w}^{t_k} p(t)dt$
where $T_s=20$ ms and $T_w=100$ ms following \cite{Aggarwal2008}. Finally, $\bf{x}$$(t_k)=[r_1(t_k),r_2(t_k),...r_q(t_k)]$ where there are $q$ biological neurons in the recording ($d=q$). As shown in Fig. \ref{fig:ELM} (b), we have $C=M+1$ output neurons in this case where there are $M$ movement types to be decoded. The $M$+1-th neuron is used to decode the onset time of movement. For decoding type of movement, we can directly use the method described in
the earlier section \ref{sec:elm} for $M$-class classifier to get the predicted output class at time $t_k$ as $s(t_k)=argmax_p o_p(t_k),1\leq p\leq M$.

For decoding movement onset time, we further create a binary classifier that reuses the same hidden layer but adds an extra output neuron. Similar to \cite{Aggarwal2008}, this output is trained for regression--the target is a trapezoidal fuzzy membership function which gradually rises from $0$ to $1$ representing the gradual evolution of biological neural activity. This output $o_{M+1}$ is thresholded to produce the final output $G(t_k)$ at time $t_k$ as:
\begin{equation}
G(t_k)=
\begin{cases}
1,\text{ if } o_{M+1}(t_k)>\theta\\
0,\text{ otherwise}
\end{cases}
\label{eq:movetime_classify}
\end{equation}
where $\theta$ is a threshold optimized as a hyperparameter. Moreover, to reduce spurious classification and produce a continuous output, the primary output $G(t_k)$ is processed to create $G_{track}(t_k)$ that is high only if $G$ is high for at least  $\lambda$ times over the last $\tau$ time points.
Further, to reduce false positives, another detection is prohibited for $Tr$ ms after a valid one. The final decoded output, $F(t_k)$ is obtained by a simple combination of the two classifiers as $F(t_k)=G_{track}(t_k)\times s(t_k)$.


\begin{figure}[!t]
\centerline{
\includegraphics[width=0.45\textwidth]{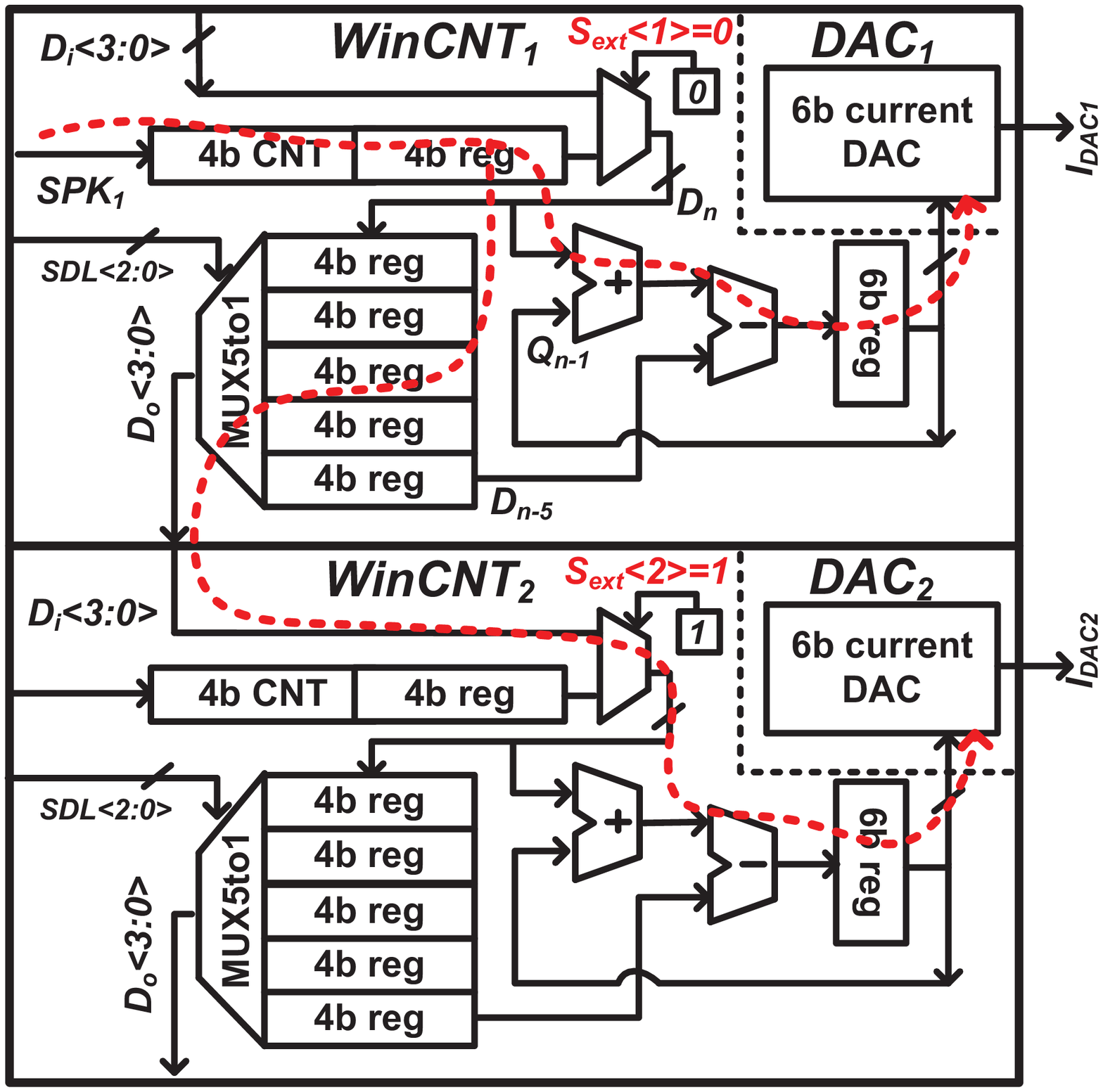}}
\centerline{(a)}

\centerline{
\includegraphics[width=0.45\textwidth]{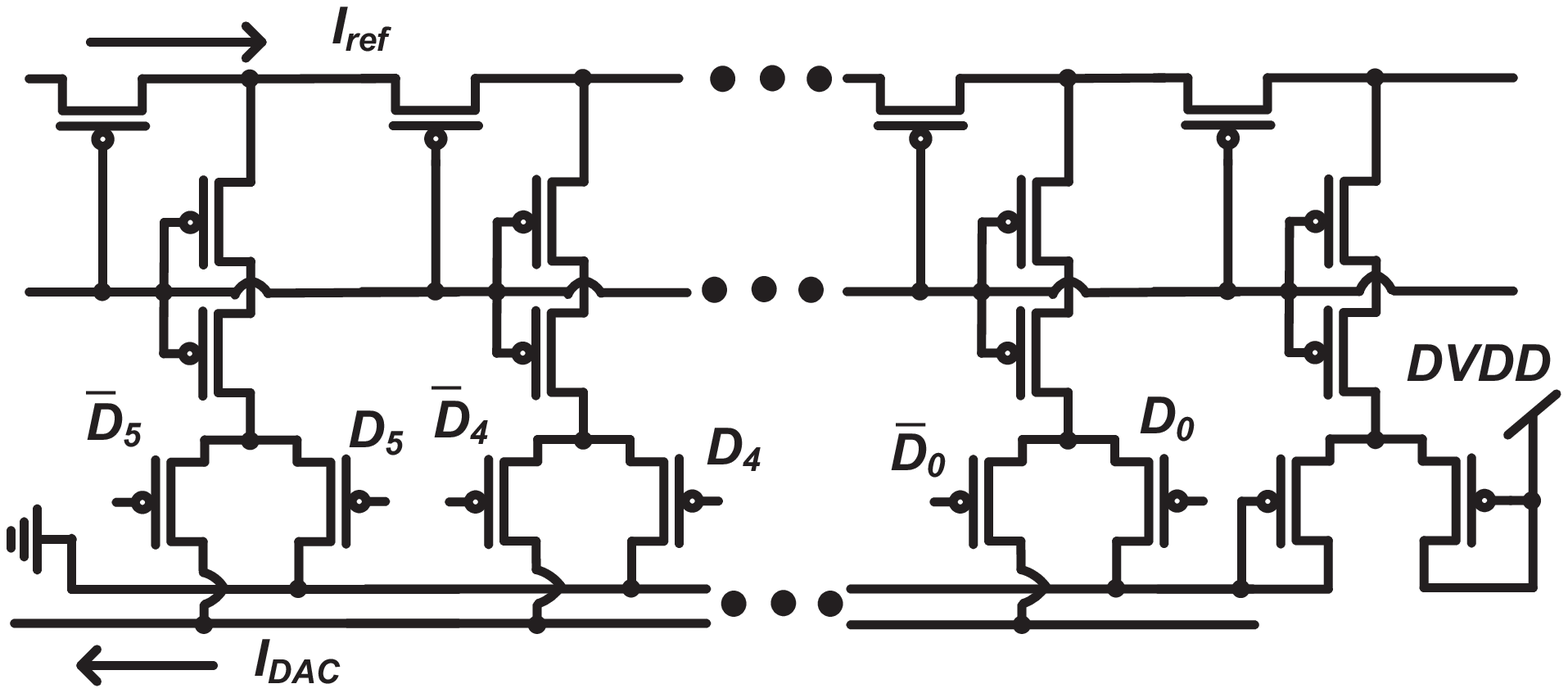}}
\centerline{(b)}

\centerline{
\includegraphics[width=0.42\textwidth]{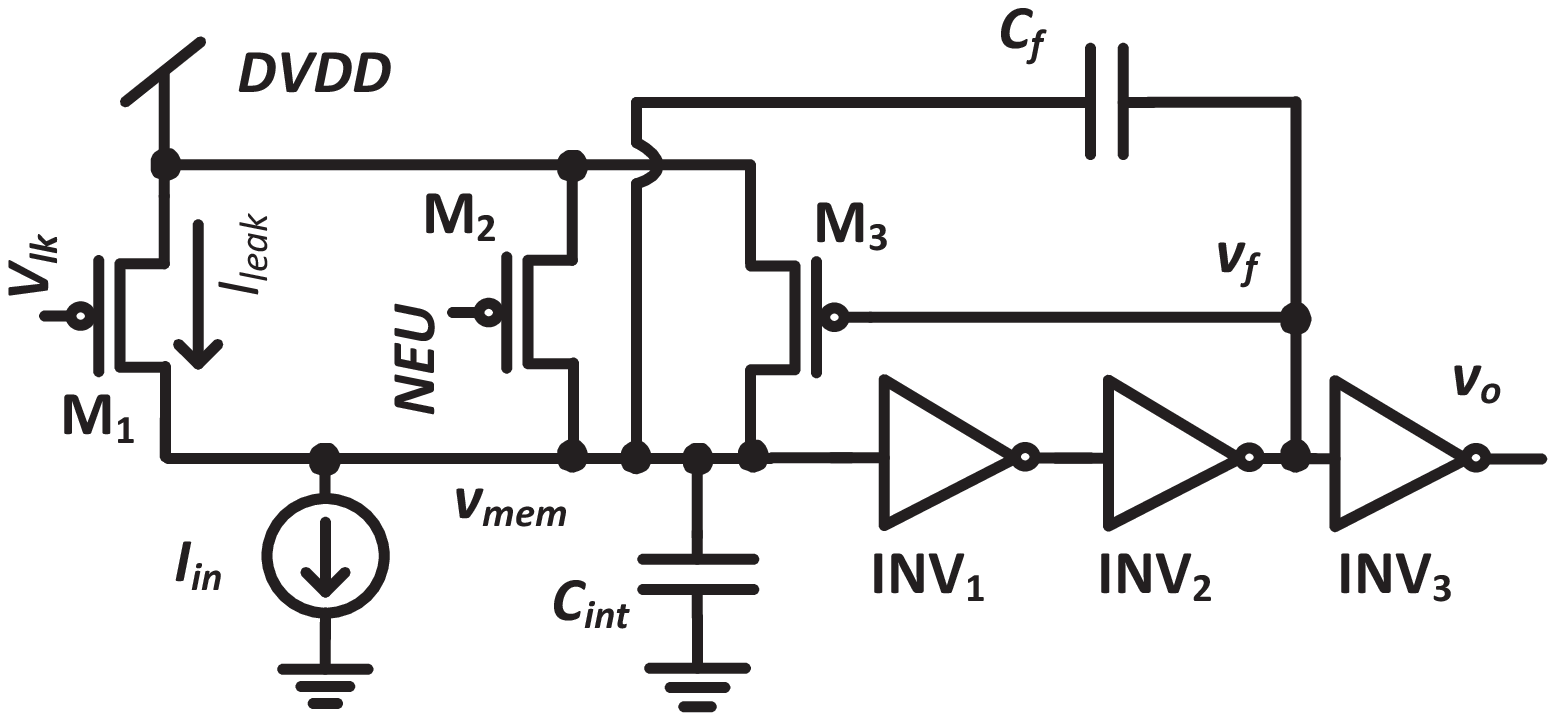}}
\centerline{(c)}
\caption[Sub-block circuit diagrams]{{\bf Sub-block circuit diagrams:} (a) Input processing circuit to take moving average of incoming spikes; (b) Current-mode DAC to convert average value to input $x$ for the current mirror; (c) Neuron-based CCO to implement hidden node non-linearity and convert to digital.}
\label{fig:circuits}
\end{figure}

\subsubsection{Time delay based dimension increase (TDBDI)}
A common problem in long-term neural recording is the loss of information from electrodes over time due to tissue reactions such as gliosis, meningitis or mechanical failures \cite{Kao2014}. Hence, initially functional electrodes may not provide information later on. To retain the quality of decoding, we propose a method commonly used in time series analysis--the use of information from earlier time points\cite{elm-timeseries}. In the context of neural decoding, it means that we use more information from the dynamics of neural activity in functional electrodes in place of lost information from the instantaneous values of activity in previously functional electrodes. So if we use $p-1$ previous values from the $n$ functional electrodes, the new feature vector is given by:
\begin{equation}
\label{eq:time_series}
\begin{split}
\textbf{x}(t_k)=[r_1(t_k),r_1(t_{k-1}),r_1(t_{k-2}),r_1(t_{k-p+1}), \\ r_2(t_k),r_2(t_{k-1})...r_n(t_{k-p+1})]
\end{split}
\end{equation}
where the input dimension of the ELM is given by $D=n\times p$. This is a novel algorithmic feature in our work compared to \cite{Aggarwal2008}. 
\section{Proposed Design: Hardware Implementation}
\label{sec:hardware}

Fig.\ref{fig:BMI} shows a typical usage scenario for our MLCP where it works in tandem with the DSP and performs the intention decoding. The DSP only needs to send very simple control signals to the MLCP and performs the calculation of the second stage of ELM (multiplication by learned weights $\beta$). The input to the MLCP comes from spike sorting that can be performed on the DSP\cite{karkare_2011}. In some cases, spike sorting may not be needed and spike detection may be the only required pre-processing \cite{Kao2014}.
\subsection{Architecture}
Details and timing of the MLCP are shown in Fig. \ref{fig:diagram}. We map input and hidden layers of ELM into the MLCP fabricated in AMS 0.35-$\mu$m CMOS process, where high computation efficiency is achieved by exploiting fabrication mismatch abundantly found in analog devices, while the output layer that requires precision and tunability (tough to attain in analog designs) can be implemented on the DSP. Since the number of computations in first stage far outnumbers those in the second (as long as $D>>C$), such a system partition still retains the power efficiency of the analog design. Up to $128$ input channels and $128$ hidden layer nodes are supported by the MLCP, with each input channel embedding an input processing circuit that extracts input feature from the incoming spike trains. As mentioned in the earlier section, we extract a moving average of the spike count as the input feature of interest.

On receiving a spike from the neural amplifier array (after spike detection and/or spike sorting), the DSP sends a pulse via $SPK$ and $7$-bit channel address (A$\left \langle 6:0\right \rangle$) to the DEMUX in the MLCP for row-decoding. Each row of the MLCP has a $6$-bit window counter ($WinCNT$) to count the total number of input spikes in a moving window with length of $5t_s$ and a moving step of $t_s$. The length of $t_s$, normally set to $20$ ms, is determined by the period of $CLK\_in$. The counter value in j-th row is converted into input feature current $I_{DACj}$ for the ELM, corresponding to the input $x_j$ in Fig. \ref{fig:ELM}. Furthermore, a $1$-bit control signal ($S_{ext}\left \langle 1\right \rangle$) stored in each row determines whether the j-th row's input to the moving window circuit is an external spike count or a delayed spike count from the previous channel. The delay length can be selected from among $5$ delay steps ranging from $20$ ms to $100$ ms, based on $SDL\left \langle2:0\right \rangle$. This is how the TDBDI feature described earlier is implemented in the MLCP.

The input feature current from each row is further mirrored into all hidden-layer nodes by a current mirror array. Hence, ratios of the current mirrors are essentially the input weights, and are inherently random due to fabrication mismatch of the transistors even when identical value is used in the design. We use sub-threshold current mirror to achieve very low power consumption, resulting in $w_{ij}=e^{\frac{\Delta V_{t,ij}}{U_{T}}}$ with $U_T$ denoting thermal voltage and $\Delta V_{t,ij}$ denoting the threshold voltage mismatch between input transistor on j-th row and mirror transistor on i-th column of that row. This is similar to the concepts described in \cite{elm-spike-mismatch}\cite{elm-biocas}. The input weights are log-normal distributed since $\Delta V_{t,ij}$ follows a normal distribution. We therefore realize random input weights in a very low `cost' way that requires only one transistor per weight. It is the fixed random input weights of the ELM that makes this unique design possible. A capacitance $C_{M}$ = $400$ fF on each row sets the SNR of the mirroring to $43$ dB.

The hidden layer node is implemented by a current controlled oscillator (CCO) driving a $14$-bit counter with a $3$-bit programmable stop value $f_{max}$ to implement a saturating nonlinearity in the activation function $g()$. The advantage of choosing this nonlinearity is that it can be digitally set and also some neurons can be configured to be linear as well to achieve good performance in linearly separable problems \cite{Miche2010}. The computation of hidden layer nodes is activated by setting $NEU$ high. The output of CCO is a pulse frequency modulated signal with the frequency proportional to total input current. The counter outputs are latched and serially read using the $CLK\_OUT$ signal when $NEU$ is low with CCO disabled to save power. The output weights, $\beta$, are stored on the DSP where the final output $o_k$ is calculated. Thus the MLCP performs the bulk of MACs (D$\times$L) while the DSP only performs C$\times$L MACs of the output layer. It should be noted here, the output of hidden layer neurons changes with power supply voltage due to sensitivity of the CCO frequency to power supply variation, leading to degradation of the decoding accuracy. However, since power supply variation is a common-mode component to all CCOs, normalization methods can be applied in post-processing (see Section \ref{sec:normalization}) to the hidden layer outputs to reduce the effect introduced by power supply variation.

\begin{figure}[!t]
	\centerline{
		\includegraphics[width=0.5\textwidth]{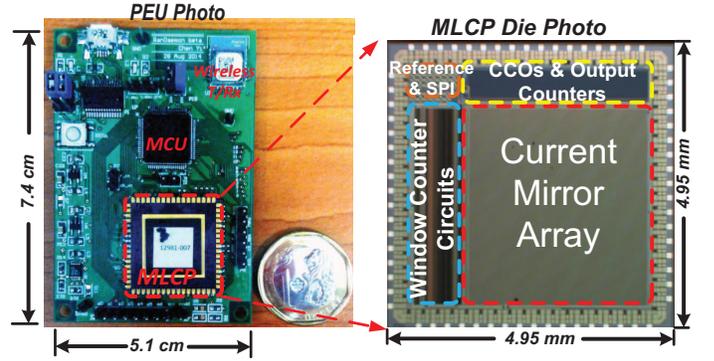}}
	\caption[Die photo and test board]{{\bf Die photo and test board:} The die photo of MLCP fabricated in $0.35$-$\mu$m CMOS process and the portable external unit (PEU) integrating MLCP with MCU and battery.}
	\label{fig:photo}
\end{figure}

\subsection{Sub-circuit: Input processing}
Fig. \ref{fig:circuits} shows diagrams of the circuit blocks in the MLCP. Fig. \ref{fig:circuits} (a) shows two adjacent input processing circuits with $WinCNT_1$ configured to receive an external spike train by setting $S_{ext}\left \langle 1\right \rangle$ = 0 and $WinCNT_2$ configured as time delay based channel by setting $S_{ext}\left \langle 2\right \rangle$ = 1. The corresponding signal flows are also depicted in the figure by red dash lines. The moving window counter is realized by (1) counting spike in a sub-window in a length of $t_s$; (2) storing sub-window counter value in a delay chain made of shift registers; and (3) adding and subtracting previous $6$-b output value with corresponding sub-window counter values in the delay chain to get new $6$-b output value of $WinCNT$. This calculation can be represented as:
\begin{eqnarray}
 Q_{n}\left \langle 5:0\right \rangle=Q_{n-1}\left \langle 5:0\right \rangle +D_{n}\left \langle 3:0\right \rangle-D_{n-5}\left \langle 3:0\right \rangle,
\label{eq:wincnt}
\end{eqnarray}
where $Q_{n}\left \langle 5:0\right \rangle$ and $D_{n}\left \langle 3:0\right \rangle$ are $6$-b output value and $4$-b sub-window counter value at time instance $n$ respectively. All registers in the input processing circuits toggle at the rising edge of $CLK\_in$. The advantage of this structure is that the delay chain for sub-window counter value is reused in the proposed TDBDI feature, leading to a compact design. 

A compact, $6$-bit MOS ladder based current mode DAC, as shown in Fig. \ref{fig:circuits} (b), splits a reference current $I_{ref}$ ($6$-bit programmable in range of $1$ nA to $63$ nA) according to the $WinCNT$ output value to generate the input feature current $I_{DAC}$ to the current mirrors.  

\begin{figure}[!t]
\centerline{
\includegraphics[width=0.45\textwidth]{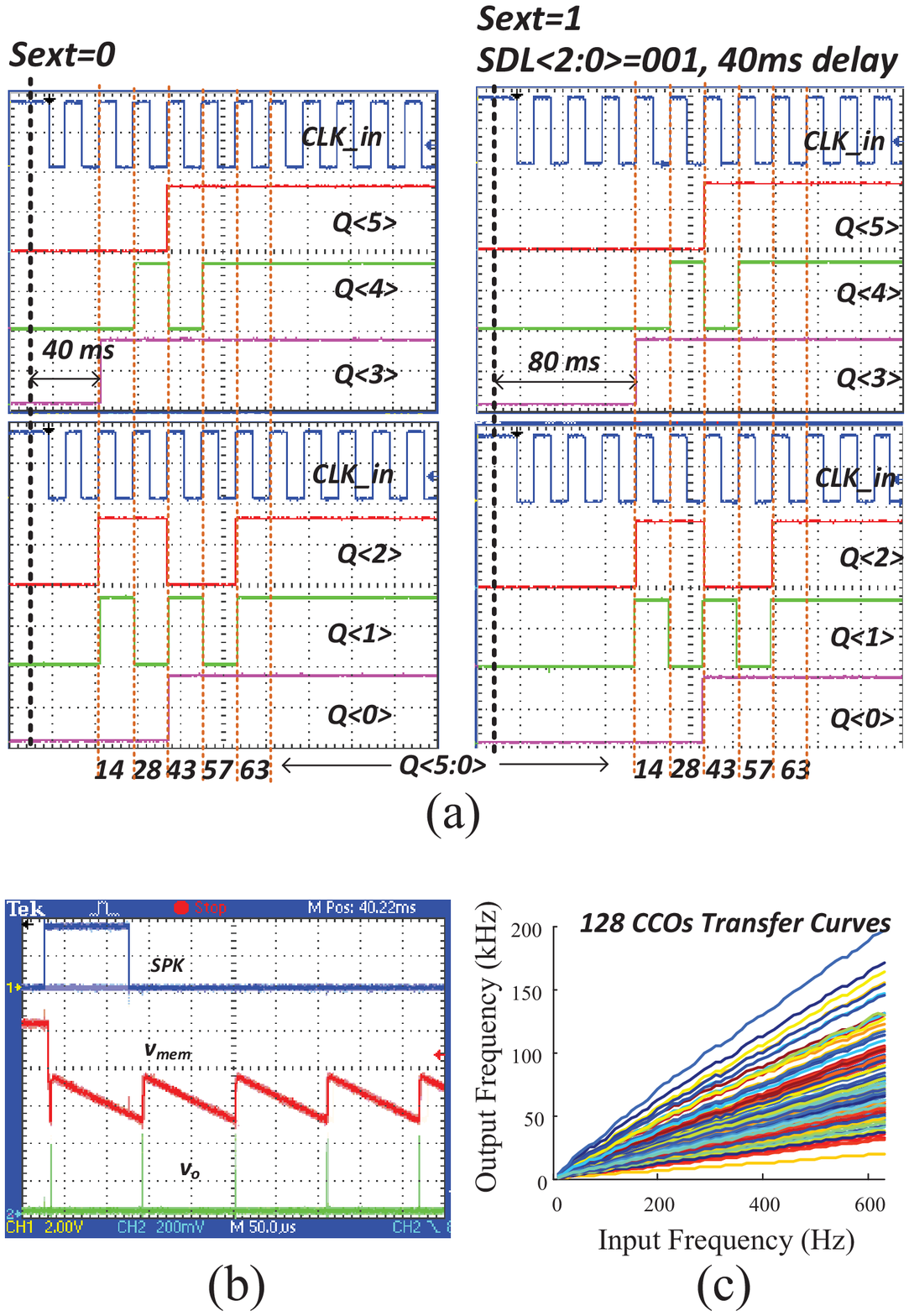}}
\caption[MLCP circuit blocks measurement results]{{\bf MLCP circuit blocks measurement results:} (a) TDBDI feature; (b) waveform of CCO oscillation; (c) transfer curves of all 128 CCOs.}
\label{fig:waveforms}
\end{figure}

\begin{figure*}[t]
	\centerline{
		\includegraphics[width=0.8\textwidth]{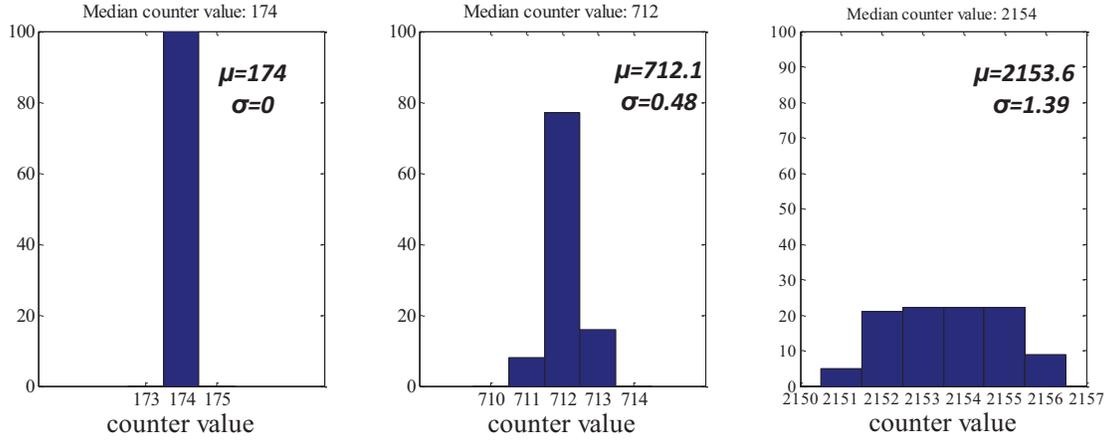}}
	\caption[Jitter performance]{\bf{Jitter performance:} The variation in the counter output for a fixed value of input current is observed for $100$ trials and plotted as a histogram for (a) low, (b) medium and (c) high input currents. The measured jitter is $<0.1\%$}
	\label{fig:jitter}
\end{figure*}

\begin{figure}[!t]
	\centerline{
		\includegraphics[width=0.45\textwidth]{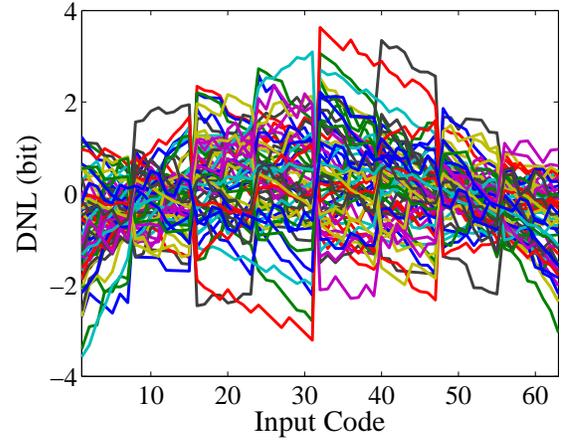}}
	\caption[DNL performance:]{{\bf DNL performance:} DNL of $64$ randomly selected input DAC channels show $\pm 3$ LSB performance.}
	\label{fig:dnl}
\end{figure}

\begin{figure}[!t]
\centerline{
\includegraphics[width=0.45\textwidth]{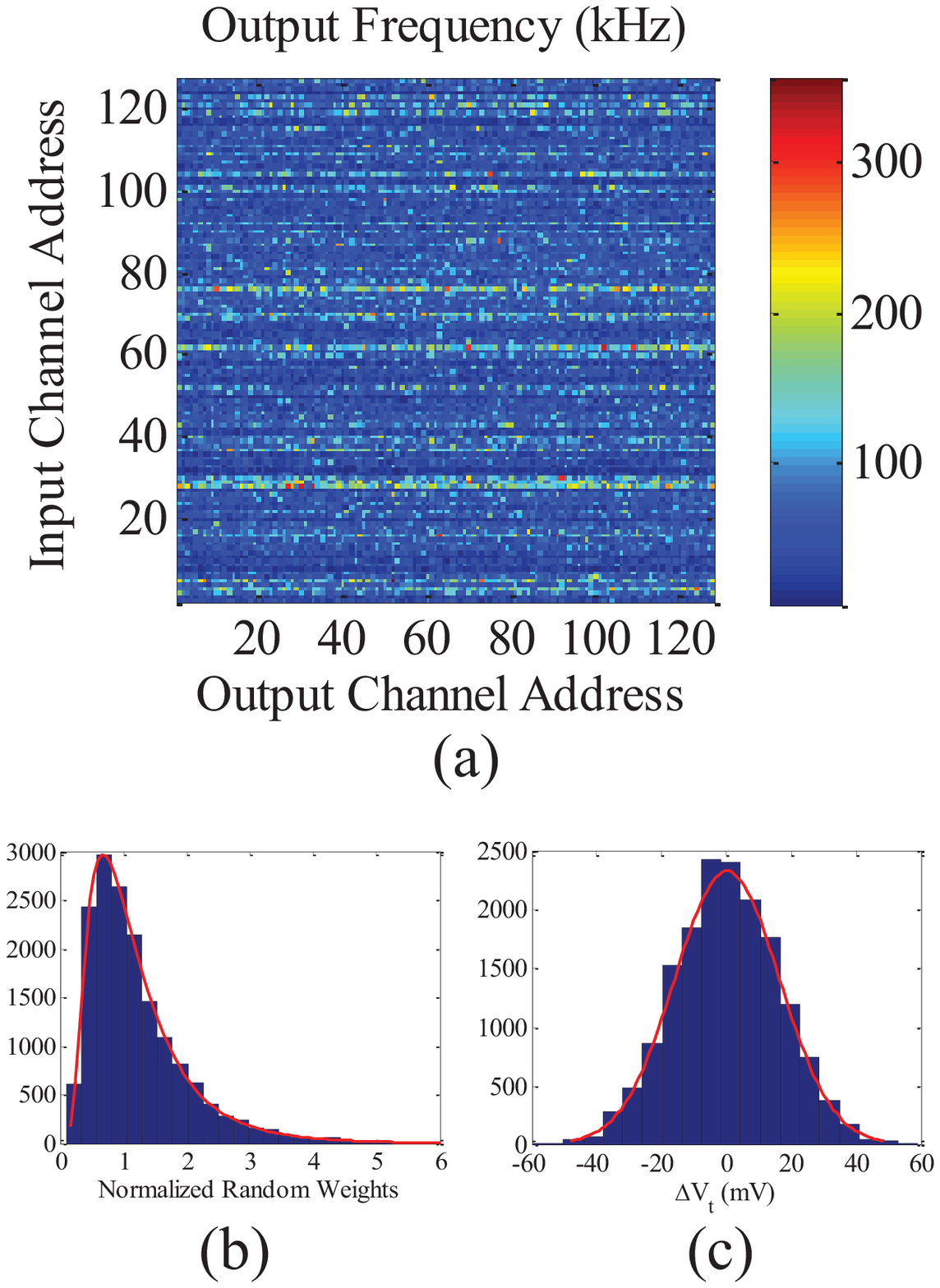}}
\caption[The random input weights]{{\bf The random input weights:} (a) Measured mismatch map of the CCO frequencies; (b) Distribution of input weights  and (c) $\Delta V_{t,ij}$. These values are measured by reading the output counter values when a fixed input value is given one row at a time.}
\label{fig:mismatch}
\end{figure}

\subsection{Sub-circuit: Current Controlled Oscillator}
The diagram of the current controlled oscillator (CCO) is depicted in Fig. \ref{fig:circuits} (c). The capacitance of $C_{int}=400$ fF sets oscillation frequency of this relaxation oscillator based on the summed input current while $C_f=100$ fF provides hysteresis through positive feedback. When $NEU$ is pulled high, pFET $\textup{M}_2$ is turned off. $M_1$ is used to set the leakage term $b_i$ in equation \ref{eq:ELM_out} and can be set to $0$ for most cases. $I_{in}$ from the current mirrors starts to discharge $v_{mem}$ until it crosses the threshold voltage of the $\textup{INV}_{1}$, leading to transition of all inverters. Then, $v_{mem}$ is pulled down very quickly through a positive feedback loop formed by $C_f$. At the same time, $\textup{M}_{3}$ turns on, charging $v_{mem}$ towards $DVDD$ until it crosses the threshold voltage of $\textup{INV}_{1}$ from low to high and the cycle repeats. Neglecting higher order effects, the time for each cycle of the CCO operation is determined by the sum of the charging and discharging time constant of $v_{mem}$, and can be expressed by:
\begin{eqnarray}
T_{CCO}=\frac{C_f\times DVDD}{I_{in}}+\frac{C_f\times DVDD}{(I_{rst}-I_{in})},
\label{eq:cco}
\end{eqnarray}
where $I_{rst}$ is the charging current when $\textup{M}_{3}$ is on. Normally $I_{rst}>>I_{in}$ reducing equation \ref{eq:cco} to: 
\begin{eqnarray}
f_{CCO}=\frac{1}{T_{CCO}}\approx\frac{I_{in}}{C_f\times DVDD}.
\label{eq:wincnt}
\end{eqnarray}


\section{Measurement Results}
\subsection{MLCP Characterization}
This section presents the measurement results from the MLCP fabricated in 0.35-$\mu$m CMOS process. To test the circuit, we have integrated it with a microcontroller unit or MCU (TI MSP430F5529) to act as the DSP. Though we have not integrated it with an implant yet, this setup does allow us to realistically assess performance of the MLCP with pre-recorded neural data as shown later. Moreover, the designed board is entirely portable with its own power supply and wireless TX-RX module (TI CC2500). Hence, it can be used as a portable external unit (PEU) for neural implant systems as well. As shown in Fig. \ref{fig:photo}, the MLCP has a die area of $4.95\times4.95$ $mm^2$ and the PEU measures $7.4$ cm $\times$ $5.1$ cm. 

\begin{table}[!t]
\begin{center}
\caption{Mean and standard deviation of $\Delta V_{t,ij}$}
\label{tab:mismatch}
\begin{tabular}{|c|c|c|}
\hline
Chip No.& $\mu$ (mV) & $\delta$ (mV) \\
\hline
1 			& 0.188			 & 16.2 \\
\hline
2 			& 0.132			 & 16.9  \\
\hline
3       & -0.019		 & 16.8 \\
\hline
4				&	-0.105		 & 17.2 \\
\hline
5				& 0.004			 & 16.5 \\
\hline
6				& 0.535			 & 16.4	\\
\hline
7				& 0.276			 & 17.6	\\
\hline 	
8				& -0.012		 & 16.6	\\
\hline
\end{tabular}
\end{center}
\end{table}

For the characterization results shown next, we use $AVDD$ = 2.1 V powering the reference circuits to generate bias currents and $DVDD$ = 0.6 V for the rest. Figure \ref{fig:waveforms}(a) verifies operation of the input processing by probing output of the window counter, with frequency of $CLK\_in$ and input spike train being $20$ Hz and $630$ Hz respectively. The output, as labeled by $Q\left \langle 5:0\right \rangle$, increases from $0$ to $63$ within $100$ ms in the left half of Fig. \ref{fig:waveforms} (a). The TDBDI feature is shown in the right half of Fig. \ref{fig:waveforms} based on setting of $SDL\left \langle2:0\right \rangle$ = 001 when $S_{ext}$ = 1. It adds a delay of $40$ ms to the $Q\left \langle 5:0\right \rangle$, comparing with waveforms in the left half. Measured charging and discharging dynamics of the CCO based neuron are shown in Fig. \ref{fig:waveforms} (b) by probing a buffered version of membrane voltage $v_{mem}$. Measured transfer curves of the $128$ CCOs in a chip is plotted in Fig. \ref{fig:waveforms} (c), by varying input spike frequency from $0$ to $630$ Hz. Here, the saturation of the count is not shown--when implemented, it stops the count at the preset value. The noise of the whole circuit is also characterized in terms of jitter at the output of the CCO. The variance in the counter value is measured for the same input current over $100$ trials. This experiment is repeated for three different current values spanning most of the counting range. The results of this experiment, shown in Fig. \ref{fig:jitter}, demonstrate percentage jitter less than $0.1\%$ for the entire counting range.

Next, we show characterization results for the input DAC channels. Since it is not possible to separately measure output current of the DAC, we measure the output of the CCO to infer the linearity of the DAC. This is reasonable since the linearity and the noise performance of the CCO is better than the $6$ bit resolution of the DAC. Figure \ref{fig:dnl} plots the measured differential non-linearity (DNL) of $64$ randomly selected input DACs. The worst case DNL is approximately $\pm 3$ LSB. While this DNL can be part of the non-linearity  $g(\bf{w_i},\bf{x},b_i)$ in the general case, it makes the implementation of the additive node less accurate.

Variation in transfer curves of the CCO array is a result of random mismatch from various aspects of the circuits, mainly current mirror array, which is expected and desired in this design. By applying the same input spike frequency of $320$ Hz to each row individually, a mismatch map of the CCO frequencies is generated with $I_{ref}=32$ nA, as presented in Fig. \ref{fig:mismatch} (a), by reading out the quantized frequency values in the output counters. These frequencies are normalized to the median frequency and plotted in Fig. \ref{fig:mismatch} (b) and (c) to show conformance to the log-normal distribution as expected. The underlying random variable of $\Delta V_{t,ij}$ has a normal distribution with mean $\approx$ 0 and standard deviation $\approx$ 16.5 mV. Totally eight sample dies are characterized with mean and standard deviation of $\Delta V_{t,ij}$ in all chips listed in Tab. \ref{tab:mismatch}.

\begin{figure*}[t]
\centerline{
\includegraphics[width=0.75\textwidth]{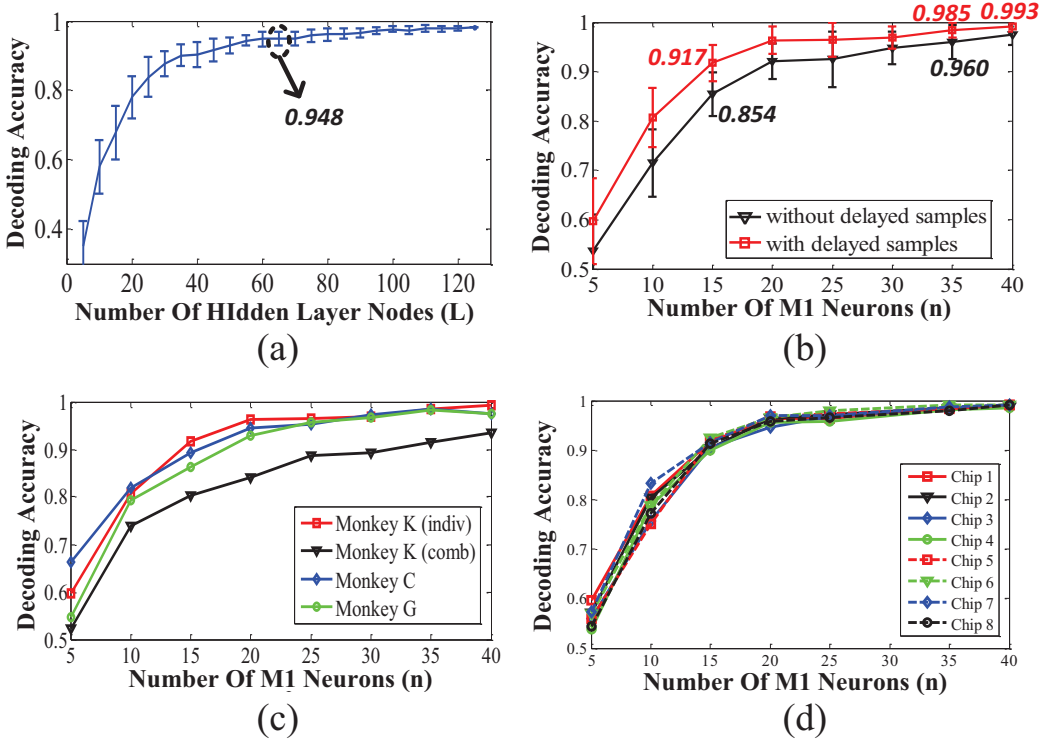}}
\caption[Measured movement types decoding performance]{{\bf Measured movement types decoding performance:} (a) Decoding accuracy versus number of hidden layer nodes; (b) Decoding accuracy versus number of M1 neurons (with/without TDBDI); (c) Decoding accuracy across monkeys; (d) Decoding accuracy across 8 dies;}
\label{fig:MC}
\end{figure*}

\subsection{Experiment}

The neural data used to verify the decoding performance of the proposed design is acquired in a monkey finger movement experiment described in detail in \cite{Aggarwal2008}. In the experiment, the monkey puts its right hand into a pistol-grip manipulandum with one finger placed in one slot of the manipulandum. The monkey is trained to perform flexion or extension of the individual finger and wrist according to given visual instruction. A single-unit recording device is implanted into the M1 cortex of the monkey, enabling real-time recording of single unit spike train during the experiment. The entire data set includes neural data recorded from three different monkeys--Monkey C, G and K, performing 12 types of individuated movements labeled by the moving finger number and by the first letter of the moving direction. Furthermore, all the trials are aligned such that the onset of the movement happens at $1$ s. Therefore the ELM can be trained according to the given label and the onset moment.

\subsection{Neural Decoding Performance}
\label{sec:decoding_meas}
We have tested the MLCP based PEU using the data set mentioned above. A multiple-output ELM with number of classes $C=12$ is trained to identify the movement type of the trial. An additional output is used to decode the onset time of movement. During training, the pre-recorded input spikes from biological neurons in M1 are sent to the MLCP the counter values of $\bf{H}$ are wirelessly transmitted to a PC where $f_{max}$ and $\beta$ are calculated and communicated back. This process already includes non-idealities in the analog processor such as DNL of input DAC, non-linearity in CCO and early effect induced current addition errors--hence, the learning takes these effects into account and corrects for them appropriately. Then, the MLCP can run autonomously during testing phase.

We present decoding results in a format similar to \cite{Aggarwal2008} for easy comparison wherever possible. For the first set of experiments, we use the normal training method T1 described in section \ref{sec:training}. As shown in Fig. \ref{fig:MC} (a) with $n=D=30$, the decoding accuracy of the $12$ types of movements (the flexion and extension of the fingers and wrist in one hand) increases as $L$ is increased, with a mean accuracy of $94.8\%$ at $L=60$. This trend is expected \cite{elm_neurocomp} since more number of random projections should allow better separation of the classes till the point when the amount of extra information for a new projection is negligible. Based on this result, we fix $L=60$ for the rest of the experiments unless stated otherwise.

\begin{figure}[!t]
	\centerline{
		\includegraphics[width=0.28\textwidth]{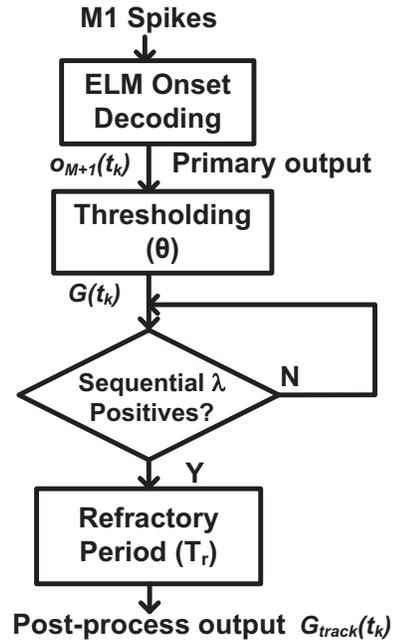}}
	\caption[Diagram of movement onset decoding]{{\bf Flow chart describing the finite state machine on DSP to calculate $G_{track}$ from $G$.}}
	\label{fig:OM_diagram}
\end{figure}

Next, we explore the variation in performance as number of available neural channels ($n$) (or equivalently M1 neurons in this case) reduces while keeping $L$ fixed at $60$. Fig. \ref{fig:MC} (b) shows that an increase in accuracy from $85.4\%$ to $91.7\%$ can be obtained at $n=15$, by using delayed samples as added features (TDBDI). Here, we have used only one earlier sample--hence, $p=2$ and the effective input dimension of the ELM is $D=2\times n$.  With $n=40$, $L=60$ and $p=2$, a decoding accuracy of $99.3\%$ can be achieved. Next, to check the robustness of the earlier result, the same experiment is performed using several different datasets, including individuated finger movement data from Monkey K, C and G and combined finger movement from Monkey K (12 individuate movements and 6 types of simultaneous movement of two fingers). The results of the MLCP with increasing M1 neurons, as shown in Fig. \ref{fig:MC} (c), is consistent with software result in \cite{Aggarwal2008}. The trend of increasing performance with more M1 neurons is expected since it provides more information. The performance of the proposed MLCP is also robust across eight sample chips, as presented in Fig. \ref{fig:MC} (d) for the same experiment as in the last two cases.

\begin{figure}[!t]
\centerline{
\includegraphics[width=0.48\textwidth]{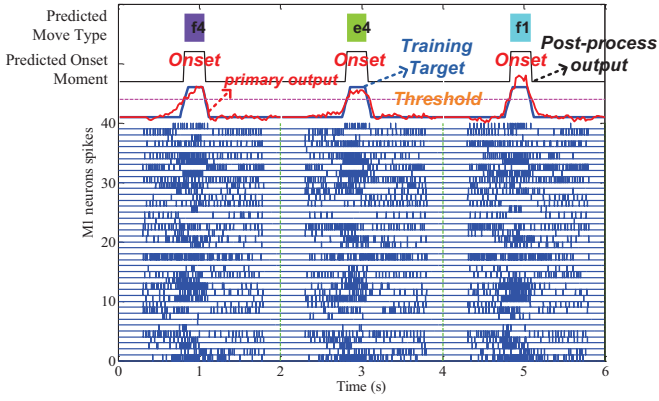}}
\centerline{(a)}
\centerline{}

\centerline{
\includegraphics[width=0.45\textwidth]{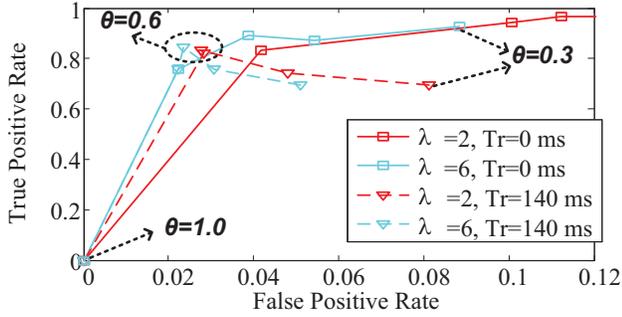}}
\centerline{(b)}
\caption[Measured movement onset decoding results]{{\bf Measured movement onset decoding results:} (a) A segment of 40 channel input spike trains is shown with real-time decoding output deciding when a movement onset happens and which tpye is this onset. (b) ROC curves of onset decoding.}
\label{fig:OM}
\end{figure}

\begin{figure}[!t]
	\centerline{
		\includegraphics[width=0.4\textwidth]{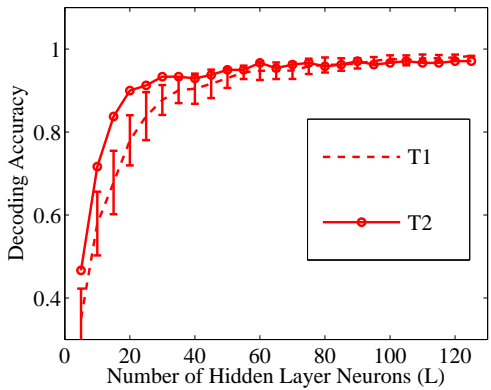}}
	\caption[Advantage of Sparsity promoting training T2]{{\bf Advantage of Sparsity promoting training T2:} The sparsity promoting method chooses best random projections and can reduce required number of hidden neurons by around $50\%$.}
	\label{fig:T2}
\end{figure}

The hidden layer output matrix $\bf{H}$ is reused to decode the onset time of finger movement using the regression capacity of the ELM. As mentioned earlier, only one more output node is added to the ELM. The trapezoidal membership function described in section \ref{sec:decoding} and shown in Fig. \ref{fig:OM} (a) is set to $1$ around the time of $1$ s to indicate the onset and set to $0$ where there is definitely no movement. Figure \ref{fig:OM_diagram} illustrates the  finite state machine in the MCU to implement the post-processing described in section \ref{sec:decoding} to obtain $G_{track}$ from the primary output $G$. Optimal values of $\lambda=6$ and $Tr=140$ ms can be found from the ROC curve shown in Fig. \ref{fig:OM} (b). The nature of the ROC curves are again very similar to the ones in \cite{Aggarwal2008}. With $\bf{H}$ reused, we achieve real-time combined decoding by detecting when there is a movement in the trial and labeling the predicted movement type when a movement onset is detected. This is illustrated by a snapshot of the developed GUI in Fig. \ref{fig:OM} (a), where three $2$-s trials are shown with $40$-channel input spike trains recorded from M1 region printed at the bottom part of the figure. Primary, post-processed output and predicted movement type are also shown in the top half of the figure.
Lastly we show the benefits of the sparsity promoting training method, T2 described in section \ref{sec:training}. To show the benefit of this method, we compare with the first experiment shown earlier in Fig. \ref{fig:MC} (a) where $n=D=30$ and the number of hidden layer neurons $L$ is varied to see its effect on performance. It can be seen that for the method T2, the decoding accuracy increases  to approximately the maximum value of $94.8\%$ attained by the method T1 for much fewer number of hidden layer neurons ($L\approx 30$). This is possible because the sparsity promoting step of minimizing $L_1$ norm of output weights chooses the most relevant random projections in the hidden layer. Thus, the new method T2 can reduce power dissipation by approximately $50\%$ due to reduction in number of hidden layer neurons.

\begin{figure}[!t]
	\centerline{
		\includegraphics[width=0.3\textwidth]{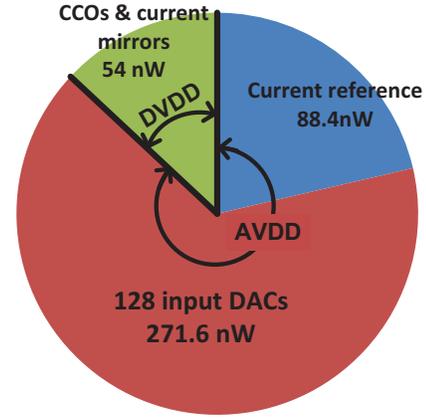}
	}  \caption[Power breakup]{{\bf Power breakup:} Power dissipation in the MLCP is dominated by fixed analog power consumption of $360$ nW compared to the power of $54$ nW dissipated from $DVDD$ in CCO and counter.} \label{fig:break_power}
\end{figure}

\begin{table*}
\begin{center}
  \begin{threeparttable}[b]
\caption{Comparison Table}
\label{tab:compare}
\begin{tabular}{|c|c|c|c|c|c|}
\hline
		& JSSC 2013\cite{Lee2013} & JSSC 2007\cite{Chakrabartty2007} & JSSC 2013\cite{Oh2013}& ISSCC 2014\cite{Lu2014} & This Work \\
\hline
Technology& 0.13 $\mu$m	 & 0.5 $\mu$m 	& 0.13 $\mu$m			& 0.13 $\mu$m & 0.35 $\mu$m	\\
\hline
Supply voltage	 & 0.85 V		& 4 V		&1.2 V (digital)& 3 V		& 0.6 V (digital)\\
								 & 					&				&1 V (analog)		&				& 1.2 V (analog) \\
\hline
Design style	&	Digital			& Analog floating 	& Mixed mode 		& Analog floating  &Mixed mode\\
							& 						& gate							&								& gate						 &					\\	
\hline
Algorithm			&	SVM			 		& SVM				& Fuzzy logic						& Deep learning feature	&	ELM feature with TDBDI \\
\hline
Application & EEG/ECG analysis	&	Speech Recognition		& Image processing	& Autonomous sensing	& Neural Implant			\\
\hline
Power	dissipation & 136.5 $\mu$W	&	0.84 $\mu$W		& 57 mW		& 11.4 $\mu$W	& 0.4 $\mu$W			\\
\hline
Max input	dimension		&	400\tnote{1}				&	14				&	14			&	8					&	128\tnote{1}		\\
\hline
Energy efficiency	&	631 pJ/MAC\tnote{2}	& 0.8 pJ/MAC	&	1.4 pJ/MAC\tnote{3}		& 1 pJ/op\tnote{4}& 5.2/1.46 pJ/MAC	\tnote{5}	\\
\hline
Resolution	& 16 b						&	4.5 b						& 6 b					& 7.5 b			& 7/14 b\tnote{6}						\\
\hline
Classification rate	& 0.5-2 Hz			& 40 Hz					& 5 MHz		& 8.3 kHz		& 50 Hz					\\		
\hline
\end{tabular}

  \begin{tablenotes}
  \item[1] can be further extended by  reusing input channels at the expense of classification rate
    \item[2] assuming 1000 support vectors
    \item[3] 1024 6-bit multiply at 10 MHz consumes 14 mW.
    \item[4] The operations are much simpler than a MAC.
    \item[5] 5.2 pJ/MAC includes both analog and digital power for $D=40$, $L=60$ and $C=12$. In reality, analog power is amortized across all multiplies and the peak efficiency of $1.46$ pJ/MAC is attainable for $D=L=128$ for the same value of $C$. See section \ref{sec:power} for details.
    \item[6] Each multiply is 7 bit accurate due to SNR limitation while the output quantization in the CCO-ADC has 14 bits for dynamic range.
  \end{tablenotes}
\end{threeparttable}  
\end{center}
\end{table*}

\subsection{Power Dissipation}
\label{sec:power}
Finally, we report the power consumption of the proposed MLCP for the $40$ input channels, $60$ hidden layer nodes, $12$-class classification problem. The current drawn from analog and digital power supply pins were measured using a Keithley picoammeter. The power breakup is shown in Fig. \ref{fig:break_power}. At the lowest value of $AVDD$ = $1.2$ V and $DVDD$ = $0.6$ V needed for robust operation, the total power dissipated is $414$ nW with $54$ nW from $DVDD$ and $360$ nW from $AVDD$.	Performing $40\times 60$ MAC in the current mirror array at $50$ Hz rate of classification, the MLCP provides a $3.45$ pJ/MAC and $8.3$ nJ/classify performance. It is clear that the efficiency is limited by the fixed analog power that is amortized across the $L$ hidden layer neurons and $D\times L$ current mirror multipliers. The fundamental limit of this architecture is the power dissipation of the CCO and current mirror array which is limited to $0.45$ pJ/MAC. 

In contrast, recently reported $16$-bit digital multipliers consume 16-70 pJ/MAC \cite{chiphong-mult}\cite{gwee-mult}\cite{razor-mult}\cite{truncated-mult} where we ignore the power consumed by the adder for simplicity. We have also implemented near threshold digital array multipliers in $65$nm CMOS operating at $0.65$ V that resulted in energy efficiency of $11$ pJ/MAC confirming the much lower energy attainable by analog solutions over digital ones. Moreover, implementing the MLCP computations in digital domain would incur further energy cost due to memory access (for getting the weight values) and clocking which are ignored here.

Since we implement the operation of second stage in digital domain, we need $C\times L$ multiplications per classification. For the case of $L=60$ and $C=12$ described above and energy cost of $11$ pJ/MAC for digital multiplies, the total energy cost of second stage operation is $7.92$ nJ/classify. Hence, the total energy/classification becomes $16.22$ nJ and the combined energy/operation increases to $5.2$ pJ/MAC. For peak energy efficiency, we consider $D=128$, $L=128$ and $C=12$ resulting in a net energy/computation of $1.46$ pJ/MAC including both stages.
\subsection{Discussion}
\label{sec:discussion}
\subsubsection{Comparison}
\label{sec:comparison}
Our MLCP is compared with other recently reported machine learning systems in Table \ref{tab:compare}. Compared to the digital implementation of SVM in \cite{Lee2013}, our implementation achieves far less energy per MAC due to the analog implementation. \cite{Chakrabartty2007},\cite{Oh2013} and \cite{Lu2014} achieve good energy efficiency similar to our method by using analog computing. \cite{Oh2013} uses a multiplying DAC (MDAC) to perform the multiplication by weights--however, they have only $6$ bit resolution in the multiply and also the MDAC occupies much larger area than the single transistor we use for multiplications. \cite{Chakrabartty2007} and \cite{Lu2014} use analog floating-gate transistors for the multiplication. Compared to these, our single transistor multiplier takes lesser area (no capacitors that are needed in floating-gates), does not require high voltages for programming charge and allows digital correction of errors because of the digital output.

\begin{figure}[!t]
	\centerline{
		\includegraphics[width=0.37\textwidth]{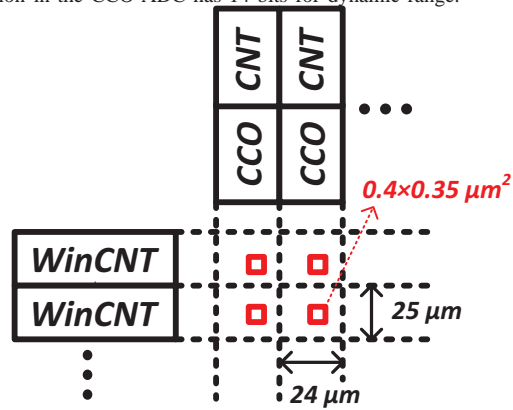}}
	\caption[Array Layout]{{\bf Array Layout:} The area of the current IC is limited by the pitch of the CCO and WINCNT circuits even though the actual area of the current mirrors ($0.4\times 0.35$ $\mu m^2$) are very small.}
	\label{fig:layout}
\end{figure}
\subsubsection{Area Limits}
Using a single transistor for multiplication in the first layer should provide area benefits over other schemes. The current layout (Fig. \ref{fig:layout}) was done due to its simple connection patterns and is not optimized. It can be seen that the actual area of a unit transistor in the array ($0.4\times 0.35$ $\mu m^2$) is much less than the area of an unit cell in the layout which is limited by the pitch of the CCO and the window counter circuits. Moving to a highly scaled process or folding the placement of the output CCO layer to be parallel to the input window counter circuits would enable large reduction ($\approx 80X$) in the area of the current mirror array. The ultimate limit in terms of area for this architecture stems from the area of capacitors--for this $128$ input, $128$ output architecture, the total capacitor area is $0.132$ mm$^2$.  
\subsubsection{Data rate requirements}
When used in an implant with offline training, the MLCP can reduce transmission data rate drastically. Firstly, for direct transmission of $100$ channel data sampled at $20$ kHz with $10$ bit resolution, required data rate is $20$ Mbps. This massive data rate can be reduced partially by including spike sorting \cite{karkare_2013}. In this case, assuming $8$ bit address encoding a maximum of $256$ biological neurons each firing at a rate $f_{bio}$, the data rate to be transmitted for a conventional implant without neural decoder is given by $R_{conv}=8\times 256\times f_{bio}$. As an example, with $f_{bio}=100$ Hz, $R_{conv}=204$ kbps. This can be reduced even further by integrating the decoder as proposed here. For the proposed case, the output of the decoder is obtained at a rate $f_{deco}$. During regular operation after training, the data rate for $C$ classes is given by $R_{prop,test}=f_{deco}\times \lceil log_2(C)\rceil$. As an example, for the case described in section \ref{sec:decoding_meas} with $f_{deco}=50$ Hz and $C=13$, $R_{prop,test}=200$ bps. This example, shows the potential for thousand fold data rate reductions over spike sorting by integrating the decoder in the implant.

From the viewpoint of power dissipation, the analog front end and spike detection can be accomplished within a power budget of $1$ $\mu$W  per channel\cite{Han2013}\cite{enyi_spd_tbcas}\cite{enyi_spd_iscas}. Assuming a transmission energy of $\approx 50$ pJ/bit from recently reported wireless transmitters for implants\cite{wireless-lian-1,wireless-minkyu-1,Chae2009}, the power dissipation for raw data rates of $200$ kbps/channel and compressed data rates of $2$ kbps/channel after spike sorting are $10$ $\mu$W and $0.1$ $\mu$W respectively. Hence, the power for wireless transmission is a bottleneck for systems transmitting raw data. For systems with spike sorting in the implant, this power dissipation is not a bottleneck. However, the power/channel needed for the spike sorter is about $5$ $\mu$W. In comparison, if our decoder operates directly on the spike detector output, it can provide compression at a power budget of $<0.01$ $\mu$W/channel. This would result in a total power dissipation/channel of $\approx 1$ $\mu$W in our case compared to $\approx 6$ $\mu$W in the case of spike sorting--a 6X reduction. There is a lot of evidence that the decoding algorithms can work on the spike detector output\cite{Kao2014}; in fact, it is believed that this will make the system more robust for long term use. This will be a subject of our future studies.

Even if the decoder is explanted, a MCU cannot provide sufficient throughput to support advanced decoding algorithms while FPGA based systems consume a large baseline power. A custom MLCP based solution provides an attractive combination of low-power and high throughput operation when paired with a flexible MCU for control flow.

\subsubsection{Normalization for Increased Robustness}
\label{sec:normalization}
\begin{figure}[!t]
	\centerline{
		\includegraphics[width=0.5\textwidth]{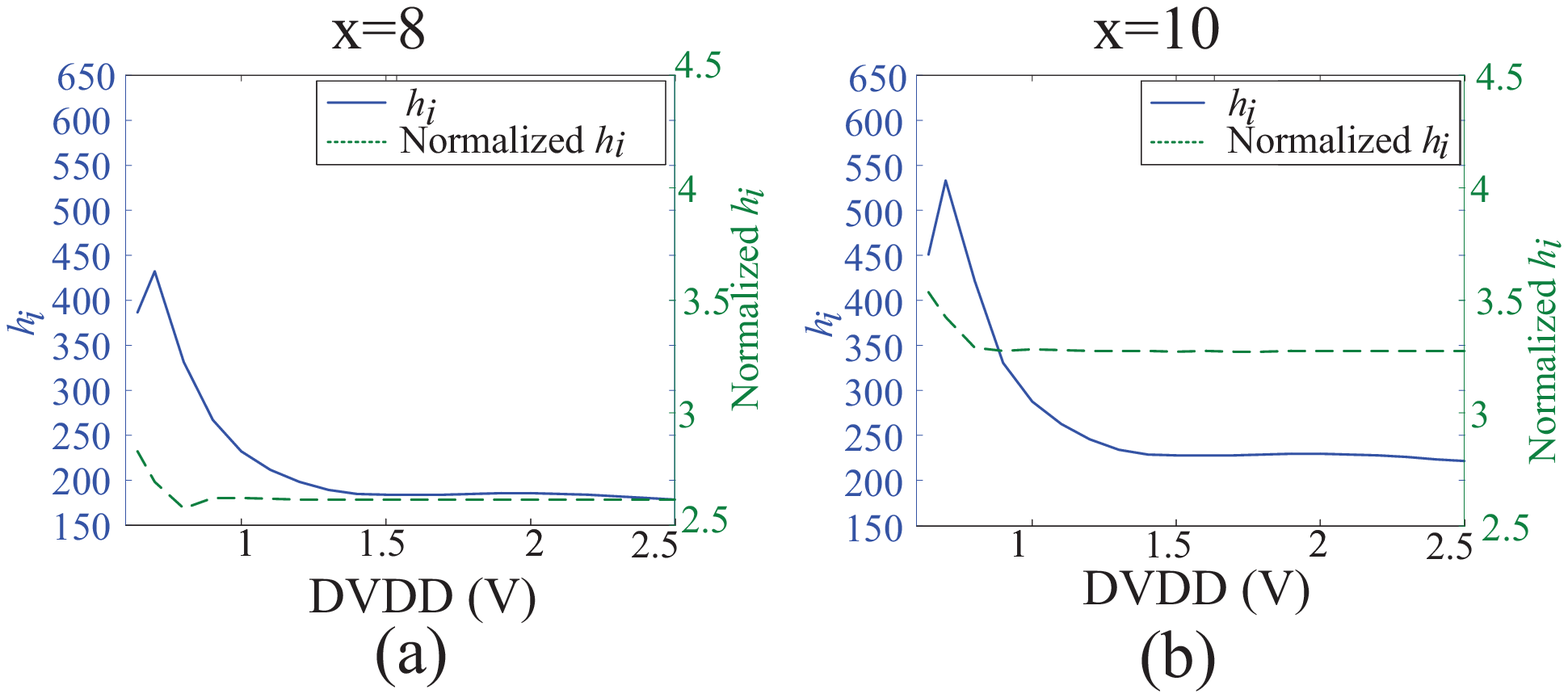}}
	\caption[Normalization to reduce variation due to $DVDD$ change]{\bf Normalization to reduce variation:} Blue lines are original hidden layer output from SPICE simulation, while green dashed lines are normalized output in both (a) and (b). The input x in (a) and (b) are 8 and 10 respectively.
	\label{fig:vdd_norm}
\end{figure}

The variation of temperature is not a big concern in the case of implantable electronics since body temperature is well regulated. However, variation of power supply voltage can be a concern. A normalization method can be applied to the hidden layer output for reducing its variation due to power supply fluctuation, at the cost of additional computation. The normalization proposed here can be expressed by:
\begin{eqnarray}
h_{j,norm}=\frac{h_j}{\sum_{j=0}^{L}h_j/\sum_{i=0}^{D}x_i}.
\label{eq:hnorm}
\end{eqnarray}
The rationale behind the proposed normalization is that the effect of power supply fluctuation on the hidden layer output can be modelled as multiplication factor in hidden layer output equation. As analyzed before, the output of the $j^{th}$ hidden layer node can be formulated as: $h_j=\frac{I_{in,j}}{C_f\times VDD}t_{cnt}$, where $I_{in,j}$ is the input current of the $j^{th}$ hidden layer node and $t_{cnt}$ is counting window length. Since $I_{in,j}$ is proportional to the strength of input vector $\textbf{x}=[x_1,x_2...x_D]$, we can model the relation between the input vector and hidden layer output as: $h_j=K_j\alpha(T,VDD)\sum_{i=0}^{D}x_i$, where the variation part is a multiplicative term $\alpha(T,VDD)$, and $K_j$ lumps up the constant part of the path gain from input to $j^{th}$ hidden layer output. It is reasonable to assume that $\alpha(T,VDD)$ is the same across different nodes, since fluctuation of power supply is a global effect on chip scale. Hence, it can be cancelled by the proposed normalization as:
\begin{eqnarray}
\begin{split}
h_{j,norm} & =\frac{h_j}{\sum_{j=0}^{L}h_j/\sum_{i=0}^{D}x_i}\\
& =\frac{K_j\alpha(T,VDD)\sum_{i=0}^{D}x_i}{\sum_{j=0}^{L}(K_j\alpha(T,VDD)\sum_{i=0}^{D}x_i)/\sum_{i=0}^{D}x_i} \\
& =\frac{K_j}{\sum_{j=0}^{L}K_j}\sum_{i=0}^{D}x_i.
\end{split}
\label{eq:cancel}
\end{eqnarray}
Simulation results are presented here to verify the proposed method of normalization. The original hidden layer outputs ($L=3$) are obtained by SPICE simulations where $DVDD$ is swept from $0.6$ V to $2.5$ V and input $x$ ($D=1$) changes from $8$ to $10$. Original and normalized values of one of the hidden layer outputs are compared in Fig. \ref{fig:vdd_norm}. As can be observed here, the normalized output (in green dashed lines) varies significantly less due to variation of $DVDD$ than the original output (in blue solid lines). The hardware cost for this normalization is $D+L$ additions and $L$ divisions. Assuming similar costs for division and multiplication, the normalization does not incur much overhead if $C>>1$ since $L\times C$ multiplications are required by the second stage anyway.

\subsubsection{Considerations for Long Term Implants}
When using this MLCP based decoder in long term implants, we have to consider issues of parameter drift over several time scales. Over long term of days, aging of the circuits in MLCP or probe impedance change due to gliosis and scarring may change performance. This is typically countered by retraining the decoder every day \cite{Kao2014}. Such retraining has allowed decoders to operate at similar level of performance over years. Over shorter time scales, any variation not sufficiently quenched by the normalization method described earlier can be explicitly calibrated by having digital multiplication of coefficients for every input and output channel. These can be determined periodically by injecting calibration inputs and observing the output of the CCO.

Another type of training--referred to as decoder retraining\cite{decoder-retrain-1,decoder-retrain-2} are needed to take into account change in neural statistics during closed loop experiments. The training done here may be thought of as open loop training for initialization of coefficents of second stage of ELM. Next, the experiment has to be redone with closed loop feedback and new training data set has to be generated for retraining the second layer weights. After several such iterations, the final set of weights of second layer will be obtained.

\section{Conclusion}

We presented a MLCP in $0.35$-$\mu$m CMOS with a die area of 4.95 $\times$ 4.95 $mm^2$ and a 7.4 cm $\times$ 5.1 cm PEU based on the proposed MLCP that achieves real-time motor intention decoding in an efficient way. Implementing the ELM algorithm, the MLCP utilizes massive parallel low power analog computing and hardware reuse, achieving a power consumption of $0.4$ $\mu$W at $50$ Hz classification rate, resulting in an energy efficiency of $3.45$ pJ/MAC. Learning in the second stage also compensates for non-idealities in the analog processor. Furthermore, It includes time-delayed sample based dimension increase feature for enhancing decoding performance when number of recorded neurons are limited. A sparsity promoting training method is shown to reduce the number of hidden layer neurons and output weights by $\approx 50\%$. We demonstrated the operation of the IC for decoding individuated finger movements using recordings of M1 neurons. However, the ELM algorithm used in the decoder is quite general and has been shown to be an universal approximator and equivalent to SVM or multi-layer perceptrons\cite{Huang2012}. Hence, our MLCP can also be used for other decoding applications requiring regression or classification computations. Higher dimensions of inputs and hidden layers can be handled by making a larger IC and also by reusing the same hidden layer several times. In either case, power dissipation increases but not energy/compute. Higher input dimensions can be accommodated at same power by reducing the bias current input of the splitter DACs in input channels\cite{elm-biocas}. Increase of hidden layer neurons however do incur a proportional power increase. Given that the power requirement of the current decoder is $>100X$ lower than the AFE, we can easily extend it to handle many more input and output channels.

\section{Acknowledgement}
The authors would like to thank Dr. Nitish Thakor for providing neural recording data.




\end{document}